\documentclass{article}

\usepackage{microtype}
\usepackage{graphicx}
\usepackage{subfigure}
\usepackage{booktabs} 
\usepackage{bm}
\usepackage{amsmath}
\usepackage{array}
\usepackage{caption}

\usepackage{hyperref}

\usepackage[accepted]{icml2019_2}

\begin{document}

\twocolumn[
\icmltitle{Maximizing Efficiency of Language Model Pre-training \\ for Learning Representation}



\icmlsetsymbol{equal}{*}

\begin{icmlauthorlist}
\icmlauthor{Junmo Kang}{equal,kaist}
\icmlauthor{Suwon Shin}{equal,kaist}
\icmlauthor{Jeonghwan Kim}{equal,kaist}
\icmlauthor{Jaeyoung Jo}{equal,kaist}
\icmlauthor{Sung-Hyon Myaeng}{kaist}
\end{icmlauthorlist}

\icmlaffiliation{kaist}{School of Computing, KAIST, Daejeon, Republic of Korea}

\icmlcorrespondingauthor{Sung-Hyon Myaeng}{myaeng@kaist.ac.kr}

\icmlkeywords{Machine Learning, ICML}

\vskip 0.3in
]



\printAffiliationsAndNotice{\icmlEqualContribution} 

\begin{abstract}
Pre-trained language models in the past years have shown exponential growth in model parameters and compute time. ELECTRA is a novel approach for improving the compute efficiency of pre-trained language models (e.g. BERT) based on masked language modeling (MLM) by addressing the sample inefficiency problem with the replaced token detection (RTD) task. Our work proposes \textit{adaptive early exit} strategy to maximize the efficiency of the pre-training process by relieving the model's subsequent layers of the need to process latent features by leveraging earlier layer representations. 
Moreover, we evaluate an initial approach to the problem that has not succeeded in maintaining the accuracy of the model while showing a promising compute efficiency by thoroughly investigating the necessity of the generator module of ELECTRA.
\end{abstract}

\section{Introduction}
\label{submission}

With the advent of self-attention based Transformer models (e.g. BERT \cite{devlin-etal-2019-bert}, XLNet \cite{NIPS2019_8812}), pre-trained language models have become the de facto standard in the current field of NLP. They form contextualized embedding for each word in a given sequence and improve the downstream tasks' performance to the state-of-the-art level. Despite the indisputable success of these models, they suffer from the problem of needing unnecessarily large compute power. This phenomenon is partly due to the masked language modeling (MLM) task proposed by BERT, in which only 15\% of the input are masked to be fed into the model and the model is trained to recover the original input words from the given context. This pretext task for pre-training a language model is greatly sample inefficient and incurs a substantial compute cost, causing the model to learn from only 15\% of the tokens per example.

ELECTRA \cite{Clark2020ELECTRA:} proposes replaced token detection (RTD), which is a token-level pre-training task that jointly trains the generator and discriminator to predict which word in a given input sequence is replaced by a plausible word; which is generated by the generator module. The benefits of RTD task are straightforward: it enables the model to learn from every token of input sequence and resolves the input mismatch problem that occurs in MLM from replacing words with a synthetic [MASK] token. The model architecture in ELECTRA is composed of two discrete modules: the generator and discriminator. Similar to BERT, the generator is trained with MLM to predict the original identities of the masked tokens in an input sequence; the generator synthesizes a plausible word in the masked word's position. Then, the corrupted sequence from the generator's output is used as the input to the discriminator, which conducts a token-wise binary classification to predict whether a given token in a sequence has been "replaced" or not (i.e. "original"). As a result, ELECTRA outperforms existing methods given the same amount of computing and uses only 25\% of the resource used to pre-train RoBERTa \cite{liu2019roberta} and XLNet to achieve comparable performance.

While ELECTRA proposes a sample efficient way of discriminating replaced tokens from the original tokens to increase the model's compute efficiency, the smaller version of the model (i.e. ELECTRA-small) still requires 4 to 5 days on a single V100 GPU to be fully trained. We present a method to maximize the training efficiency of ELECTRA by adaptively applying the early exit \cite{schwartz-etal-2020-right} strategy to the generator. Our method is predicated on the intuition that every instance varies in the level of difficulty. For example, an input sequence "I like apple," should be relatively easier to encode than "Health experts say the spikes in new cases now coincide with states starting to reopen several weeks ago." To elaborate, this idea proposes that easier instances can be confidently encoded by the earlier layers of the model whereas harder instances should propagate through the subsequent layers to be encoded into good feature representations. Based on this idea, we hypothesize that samples from earlier layers of the generator should be easier for the discriminator (i.e. higher accuracy), and the samples from later layers of the generator should be more difficult for the discriminator (i.e. lower accuracy).

Our work imposes probabilistic adaptive early exit on our generator. Our approach allows for the discriminator to determine which layer of the generator should be sampled to create a suitable input to the discriminator. Such layer-wise probability for selecting a layer depends on the performance of the discriminator (i.e. accuracy) and this adaptive early exit strategy is explained in more detail in Section 2. Furthermore, we also evaluate an experimental approach, which is to replace the role of generator with top-$k$ similar word search for improved compute efficiency. The evaluation result shows that the role of the generator in providing a challenging, meaningfully confusing input to the discriminator is necessary to train a language modeling discriminator.


\section{Approach}

\subsection*{Generator Replacement}
The increased compute time, memory consumption caused by the generator and its simple objective of replacing input tokens to the discriminator with probable words made us question the validity and usefulness of the generator. We assumed that we could either (i) add random noise to the token-to-be-replaced, rather than actually replacing it, or (ii) choose the nearest top-k words in the embedding space and use them to replace 15\% of the discriminator input sequence.
\begin{figure}[h]
    \centering
    \includegraphics[width=0.5\textwidth]{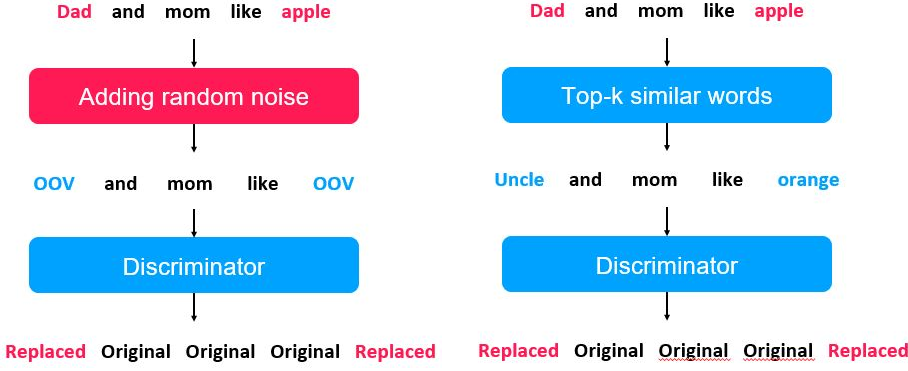}
    \caption{Left: Noise addition, Right: Top-k similar words}
    \label{fig:gen_replace}
\end{figure}
\raggedbottom

The first approach of adding random noise to the 15\% of the discriminator inputs turned out much easier for the discriminator. We added a noise sampled from normal distribution to the embeddings corresponding to the replaced tokens. After only 100 iterations, the discriminator achieves 100\% accuracy on the noised embeddings. This suggests that simply adding a random noise creates out-of-vocabulary (OOV) embeddings, allowing easy binary classification task for the discriminator to solve. 

Then, we decided to choose the nearest words in the embedding space (i.e. top-$k$) using a simple L2 distance based similarity function. Also, we added additional loss term which minimizes the L2 distance between the original and replaced token embedding vectors in order to offer stability. This process not only eliminates the need to train an entire generator with MLM for making a plausible-word prediction, but it also solves the class imbalance problem. In the previous architecture (i.e. generator and discriminator), as the generator progressively learns and MLM accuracy increases, the input to the discriminator gradually becomes dominated by the original tokens - this creates the class imbalance problem. If we remove the generator and replace the 15\% of the tokens with top-$k$ similar words, we can maintain the ratio of replaced tokens within our sequence.

\subsection*{Adaptive Early Exit}

Early exit \cite{schwartz-etal-2020-right} is a recently proposed efficient computation strategy for BERT. The authors of Early exit claim that lower layers of BERT is enough to process easy instances and thus do not need to propagate all the way up to the later layers. This halting-in-the-middle strategy improves the compute efficiency by reducing the number of FLOPs required to propagate an instance. Based on this idea, we propose the \emph{adaptive early exit} for the generator.

We take two separate approaches to test the viability of our adaptive early exit strategy:
\begin{enumerate}
   \item Applying a regular early exit to the discriminator.
   \item Applying adaptive early exit to the generator.
\end{enumerate}

\begin{figure}[h]
    \centering
    \includegraphics[width=0.5\textwidth]{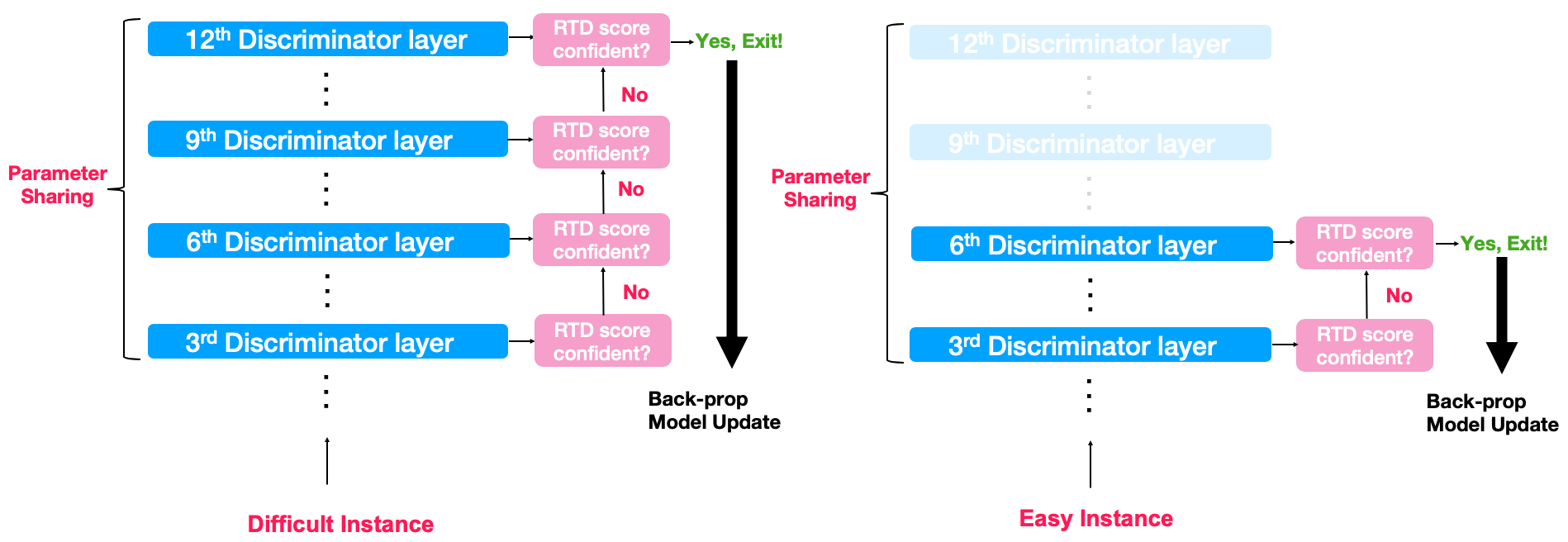}
    \caption{Early-Exit to Discriminator}
    \label{fig:early_exit(Disc)}
\end{figure}

For the first approach as shown in Figure \ref{fig:early_exit(Disc)}, we divide the 12 layers of the discriminator module into 4 sections: layers 1 to 3, 4 to 6, 7 to 9, and 10 to 12. Each section is represented by its largest layer number (i.e. layers 1 to 3 would be layer 3), and every 100 iterations we check if the section-level RTD accuracy exceeds the set threshold, which is the confidence score to make the early exit decisions. We calculate the threshold as the average of the RTD accuracy from the 4 sections and update its value every 100 iterations. We also share the parameters of the generator and discriminator to prevent later layers of the generator from being under-trained in the process.

\raggedbottom

\begin{figure}[h]
    \centering
    \includegraphics[width=0.5\textwidth]{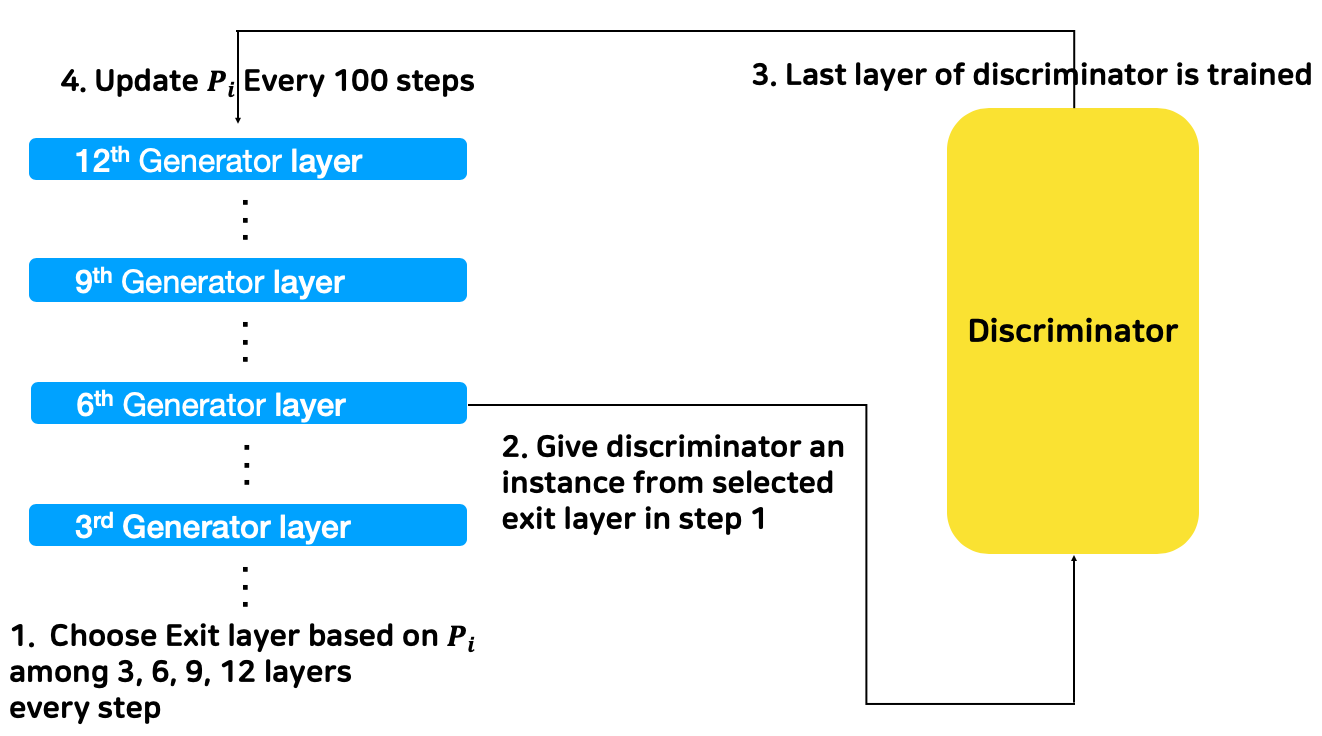}
    \caption{Early-Exit to Generator}
    \label{fig:early_exit(Gen)}
\end{figure}

With the second approach of applying the \emph{adaptive early exit} to the generator shown in Figure \ref{fig:early_exit(Gen)}, we assign a probability $P_{i} = [p_{3}, p_{6}, p_{9}, p_{12}]$ to each section (i.e. layers 3, 6, 9, 12) of the generator, with the initial probability of $P_{1}$. These probability values change according to $diff_i$, which is the difference between the averaged RTD accuracy over all the sections of the previous and current 100 iterations. Unlike the previous approach, the inputs fully propagate through the discriminator while using the generator adaptively with the given probability settings. The update process of $P_{i}$ is
\begin{align*}
    P_{1} &= [0.1, 0.2, 0.3, 0.4]\\
    R &= [0, 1, 2, 3]\\
    diff_{i} &= acc_{i} - acc_{i-1}\\
    diff_{i} &= min(1, max(-1, diff_{i}))\\
    S_{i} &= \alpha * diff_{i} * R\\
    P_{i} &= Softmax(P_{i-1} + S_{i}) 
\end{align*}
where $R$ is the \emph{reassignment score} constants for reallocating probabilities to control the difficulty of discriminator input sequence and $\alpha$ is a hyperparameter that controls scale of the score or decides sharpness of probabilities. To be specific, $R$ increases the probability of selecting a generator layer that makes the task easier for the discriminator if the previous RTD accuracy is lower than the current RTD accuracy and vice versa. For example, if $P_{k-1} = [0.4, 0.4, 0.1, 0.1]$ at step $k$ with $acc_k = 0.9$ and $acc_{k-1} = 0.7$, then following the above update process gives $P_k = [0.27, 0.28, 0.21, 0.21]$, thus increasing the likelihood of selecting more "challenging" replacements for the discriminator to classify. This creates a constant shift between the easier instances and the more difficult instances according to the change in the RTD accuracy.

The initial assignment of $R \in [0, 3]$ is based on the idea of monotonicity. We assume monotonicity at the beginning of the pre-training step, because we hypothesize that instances from the lower layers of the generator are easier for the discriminator to handle and thus gives high RTD accuracy, and the instances from the higher layers of the generator are harder for the discriminator and gives low RTD accuracy. Under such assumption, we selectively control the discriminator input difficulty.

\begin{figure}[h]
    \centering
    \includegraphics[width=0.5\textwidth]{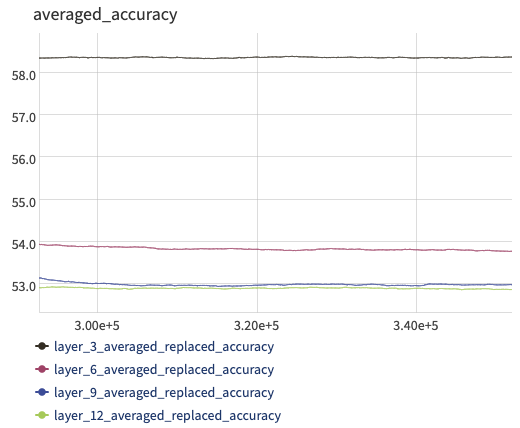}
    \caption{RTD accuracy of each exit layer}
    \label{fig:monotonicity}
\end{figure}

We use two strategies to encourage monotonicity in generator. We give more information that is concatenation of previous layers to upper layer as input to upper layer and give more weight to loss of upper layer. By these strategies, upper layer of generator is trained as optimal layer than other layers, so it results in that RTD accuracy of instance from lower layer of generator is higher than instance from upper layer. As we can see in Figure \ref{fig:monotonicity}, monotonicity seems to be well kept.

\raggedbottom

\section{Experiments}

\subsection*{Dataset}
\begin{figure}[h]
    \centering
    \includegraphics[width=0.5\textwidth]{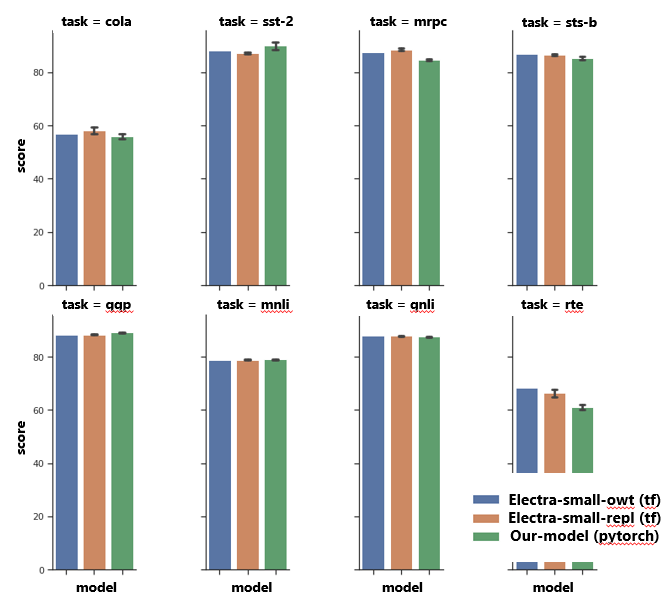}
    \caption{GLUE benchmark evaluation on the replication}
    \label{fig:GLUE_replication}
\end{figure}
\raggedbottom

We use the OpenWebTextCorpus (OWT) \cite{Gokaslan2019OpenWeb}, a 12GB raw text English dataset, to pre-train our language model. With the absence of publicly available BookCorpus data, we could not use the same dataset previously used to pre-train ELECTRA-small in the original paper. After tokenizing and encoding sequence instances into Lightning Memory-mapped Database (LMDB) using \texttt{transformers} and \texttt{pyxis} library, we load the dataset without further preprocessing to reduce on-the-fly processing time during pre-training step. For evaluation, we used the General Language Understanding Evaluation (GLUE) benchmark to test our model's performance on eight different NLP tasks.

\subsection*{Experimental Setup}
Our experiments for pre-training ELECTRA are conducted with V100 and RTX 2080Ti GPUs. For the pre-training task with batch size equal to 128 (same with official ELECTRA-small), we use V100 GPU to train our model for 5 days. With adaptive early exit, we add output layers to the following four layers in the generator: 3, 6, 9 and 12. The decision to select from one of these four layers is determined by the adaptive probability, which is updated every 100 steps during training. 

For comparative analysis in our replication process, all of the models are trained for 1M steps each. However, to compare the results of our improvement approach (i.e. adaptive early exit and generator replacement) in a very limited time constraint, we compare the evaluation results on the pre-trained ELECTRA-small (our model) baseline trained for 200K steps.

We compare three different scores based on the GLUE benchmark: (i) ELECTRA-small-OWT, which is the reported GLUE benchmark scores on the official Google Research Github page, (ii) ELECTRA-small-repl., which is the official implementation in Tensorflow by Google Research, but pre-trained in our environment, and (iii) our model, which is a PyTorch implementation of ELECTRA-small (Figure \ref{fig:GLUE_replication}).

Our replicated version of ELECTRA-small performs on par with the two official scores as shown in Figure \ref{fig:GLUE_replication}. This proves that our implementation of ELECTRA is successful and can further be used in future research tasks that build upon this model. Although some tasks (e.g. QNLI, RTE) appear to perform slightly worse than the official implementation, such differences are negligible due to big variance between each fine-tuning trials.

\begin{table*}[t!]
\begin{center}
\begin{tabular*}{\linewidth}{lccccccccc}
\toprule
\textbf{Model}     & \textbf{\small{CoLA}} & \textbf{\small{SST-2}} & \textbf{\small{MRPC}} & \textbf{\small{STS-B}} & \textbf{\small{QQP}} & \textbf{\small{MNLI}} &     \textbf{\small{QNLI}} & \textbf{\small{RTE}} & \textbf{\small{Steps/Sec}} \\ \midrule\midrule
Baseline  &   52.40   &   87.04   &   82.84   &   84.11   &   88.04   &   76.09   &   85.53   &   58.12   &   2.4 (1x) \\ \midrule
Emb-Gen  &   15.61   &   82.68   &   68.62   &   73.68   &   86.28   &   69.83   &   81.18   &   51.62  &   4.6 (1.92x)\\ 
Emb-Gen (Pre-trained)  &   24.80   &   86.00   &   71.07   &   77.88   &   86.47   &   72.78   &   82.31   &   50.18   &   4.6 (1.92x) \\ \midrule
Early-Exit (Disc)  &   40.22   &   81.99   &   83.08   &   81.19   &   86.70   &   72.74   &   83.89   &   61.01 &   3.1 (1.29x)\\ 
\textbf{Adaptive Early-Exit (Gen)}      & 49.27 & \textbf{87.38}  & \textbf{83.82}    & 82.96    & \textbf{88.33}    & \textbf{77.39}  & 84.38    & \textbf{62.81}   &  \textbf{2.75 (1.15x)}          \\ \bottomrule
\end{tabular*}
\end{center}
\caption{GLUE benchmark evaluation on the baseline, embedded-generator, embedded-generator (pre-trained embedding), Early-Exit (Disc) and Adaptive Early-Exit (Gen). All the models are trained for 200K steps for comparison.}
\label{table:GLUE_improvement}
\end{table*}

\subsection*{Implementation Details}
Prior to experiments, we address some discrepancies between the actual implementation and the settings specified in the original work.
During the re-implementation process, we encountered two major differences in the official implementation in Tensorflow by Google Research from the model architecture specified in the original paper \cite{Clark2020ELECTRA:}. We also use Huggingface's \texttt{transformers} \cite{Wolf2019HuggingFacesTS} library as a reference to refine our replication process. The official implementation for ELECTRA, unlike the original paper, adds a noise (i.e. Gumbel noise) to the logit output of the generator before feeding it to the softmax layer for sampling as follows:
\begin{equation}
p_{G}(x_t | \bm{x}) = \frac{exp(e(x_t)^T h_G(\bm{x})_t + g)}{\sum_{x^{'}}{exp(e(x^{'})^{T} h_G(\bm{x})_t + g)}}
\end{equation}
\begin{equation}
\hat{x_t} = argmax_{k} p_{G}(x_t | \bm{x})_{k}
\end{equation}
$g$ is the noise from Gumbel distribution added to the raw logits of the generator $G$, $e$ denotes token embedding, and $\hat{x_t}$ denotes the sampled word to replace the original token at $t$ (or remains the original token) in the discriminator input sequence. The addition of noise in our assumption is to control the performance of the generator and mitigate the MLM accuracy, because if the generator does too well (i.e. high MLM accuracy), discriminator is going to have a very well-synthesized input sequence leading to increased difficulty (i.e. low RTD accuracy) in predicting which token is replaced or not. 

ELECTRA maps a sequence of input tokens $\bm{x}$ into a sequence of contextualized vector representations. For a given position $t$, the discriminator predicts whether the token $x_t$ is “real,” i.e., that it comes from the data rather than the generator distribution, with a sigmoid output layer:
\begin{equation}
D(\hat{\bm{x}}, t) = sigmoid(w^T h_D(\hat{\bm{x}})_t)
\end{equation}
where $w$ is the fully connected layer on top for binary classification and $\hat{\bm{x}}$ is the input sequence that has been corrupted by the generator-synthesized words.

Aside from adding noise to the logits, the online implementation uses different masking rates compared to BERT \cite{devlin-etal-2019-bert}. In the original BERT, the authors choose 15\% of the input sequence as the masked tokens. Out of these tokens, only 80\% of them are replaced with the actual [MASK] tokens, 10\% to random tokens, and the other 10\% remain as the original tokens. However, the official implementation replaces 85\% of the selected tokens (i.e. 15\% of the input sequence) with the [MASK] and the remaining 15\% remains as the original tokens as shown in Table \ref{table:mask_ratio}.

In BERT, 10\% of masking tokens are replaced to random tokens because diminish the discrepancy that [MASK] tokens does not appear during fine-tuning. However, ELECTRA only uses discriminator that does not use [MASK] token as input during pre-training, so random tokens are not needed anymore.

\begin{table}[h!]
\centering
\begin{tabular}{ |c|c|c| } 
\hline
   & BERT & ELECTRA \\ [0.5ex] 
\hline
[MASK] & 80\% & 85\% \\ 
\hline
Random & 10\% & 0\% \\
\hline
Original & 10\% & 15\%  \\ [1ex] 
\hline
\end{tabular}
\caption{Ratio of the masked to original tokens out of 15\% of the input sequence in BERT and ELECTRA}
\label{table:mask_ratio}
\end{table}

Overall, the generator is trained to perform MLM and the discriminator is trained to discriminate "replaced" tokens by the generator out of real, "original" tokens. We then minimize the combined loss as follows:

\begin{equation}
\min_{\theta_G, \theta_D} \sum_{x \in \chi} \mathcal{L}_{MLM}(\bm{x}, \theta_{G}) + \lambda\mathcal{L}_{Disc}(\bm{x}, \theta_D)
\end{equation}

$\chi$ is a large corpus of raw text and $\mathcal{L}_{MLM}(\bm{x}, \theta_{G})$ and $\mathcal{L}_{Disc}(\bm{x}, \theta_D)$ refer to the generator and the discriminator loss functions, respectively. As in \citet{Clark2020ELECTRA:}, we disregard the generator after pre-training and fine-tune the discriminator on downstream tasks.

To test the two approaches, we experiment with the same RTD task for the adaptive early exit and generator replacement tasks. 

\section{Results}


\subsection*{Generator Replacement}


In Table \ref{table:GLUE_improvement}, we see two top-$k$ generator replacement models (i.e. Emb-Gen and Emb-Gen (Pre-trained)). Emb-Gen is a short for "Embedding Generator," which searches the word embedding space to compute similarity scores with the replaced words and substitutes the similar words in place of those words. This approach not only eliminates the need of a generator module but also accelerates the training speed by 92\% at 4.6 steps per second. Nevertheless, the speed gain from removing the generator results in huge performance drop for both models. Such result implies that the generator's job is more complex than simply finding similar words and passing them to the discriminator.

Emb-Gen (Pre-trained) is a model that uses the embeddings from ELECTRA-small baseline model that was trained for 1M steps. We include the results for Emb-Gen (Pre-trained) to show that learning from scratch with untrained embeddings as in the raw Emb-Gen is a very difficult task and requires additional contextual-level or word-level information to reach the baseline performance.

\subsection*{Adaptive Early Exit}

For the adaptive early exit task, we name the first approach of applying early exit on discriminator only as \emph{Early-exit (Disc)} and the second approach of applying adaptive early exit on generator only as \emph{Adaptive Early-exit (Gen)}. Table \ref{table:GLUE_improvement} shows the results of these two approaches against the ELECTRA-small baseline trained for 200K steps.

With Early-Exit (Disc), we identified the insufficient learning occurring in the later layers due to premature exits in the middle layers preventing the training instances from fully propagating through all the layers of the discriminator. Although applying early exit only in the discriminators increases the speed of training by 30\%, it causes underfitting in the later layers during the inference stage, causing a large drop in performance as we can see in Table \ref{table:GLUE_improvement}.

From our observation of Early-Exit (Disc), we propose Adaptive Early-Exit (Gen) for adaptive early exit to not only solve the inference mismatch problem, but also to provide a direct feedback to the generator in layer selection. Adaptive Early-Exit (Gen), while increasing the training speed by 15\%, shows a comparable performance to the baseline model in GLUE evaluation tasks (Table \ref{table:GLUE_improvement}).

\section{Discussion and Conclusion}

In this work, we have worked on the problem of efficient pre-training. During the process of solving the issues related to model replication, we identified a number of problems in pre-training, which led us to come up with several ideas for improvement. Thus, several plausible approaches have been proposed, ending up with a promising result with the adaptive early exit model (i.e. Adaptive Early-Exit (Gen)).

One of the limitations of this work is the lack of sufficient number of experiments. Due to limited time and resource constraints, which are critical for pre-training research, our models are pre-trained for only 200k steps. Out of two improvement approaches we evaluated, we could expect training \emph{Adaptive Early-Exit (Gen)} for 1M steps to yield clearly better results than the baseline model. Such expectation is based on the observation that \emph{Adaptive Early-Exit (Gen)} performs better than the baseline even though the generator of our approach is not converged yet because of the enforced early-exit strategy to balance the training of discriminator. Once the generator converge (over 1M steps), our discriminator that is adaptively trained will be more robust by leveraging the generator that has full capability of controlling the difficulty.

Future work includes developing the promising approach elaborately with the insight and intuition obtained here, and extensive experiments would be required to check the validity of the aforementioned approaches.

\bibliography{ksc}

\begin{thebibliography}{7}
\providecommand{\natexlab}[1]{#1}
\providecommand{\url}[1]{\texttt{#1}}
\expandafter\ifx\csname urlstyle\endcsname\relax
  \providecommand{\doi}[1]{doi: #1}\else
  \providecommand{\doi}{doi: \begingroup \urlstyle{rm}\Url}\fi

\bibitem[Clark et~al.(2020)Clark, Luong, Le, and Manning]{Clark2020ELECTRA:}
Clark, K., Luong, M.-T., Le, Q.~V., and Manning, C.~D.
\newblock Electra: Pre-training text encoders as discriminators rather than
  generators.
\newblock In \emph{International Conference on Learning Representations}, 2020.
\newblock URL \url{https://openreview.net/forum?id=r1xMH1BtvB}.

\bibitem[Devlin et~al.(2019)Devlin, Chang, Lee, and
  Toutanova]{devlin-etal-2019-bert}
Devlin, J., Chang, M.-W., Lee, K., and Toutanova, K.
\newblock {BERT}: Pre-training of deep bidirectional transformers for language
  understanding.
\newblock In \emph{Proceedings of the 2019 Conference of the North {A}merican
  Chapter of the Association for Computational Linguistics: Human Language
  Technologies, Volume 1 (Long and Short Papers)}, pp.\  4171--4186,
  Minneapolis, Minnesota, June 2019. Association for Computational Linguistics.
\newblock \doi{10.18653/v1/N19-1423}.

\bibitem[Gokaslan \& Cohen(2019)Gokaslan and Cohen]{Gokaslan2019OpenWeb}
Gokaslan, A. and Cohen, V.
\newblock Openwebtext corpus, 2019.

\bibitem[Liu et~al.(2019)Liu, Ott, Goyal, Du, Joshi, Chen, Levy, Lewis,
  Zettlemoyer, and Stoyanov]{liu2019roberta}
Liu, Y., Ott, M., Goyal, N., Du, J., Joshi, M., Chen, D., Levy, O., Lewis, M.,
  Zettlemoyer, L., and Stoyanov, V.
\newblock Roberta: A robustly optimized bert pretraining approach, 2019.

\bibitem[Schwartz et~al.(2020)Schwartz, Stanovsky, Swayamdipta, Dodge, and
  Smith]{schwartz-etal-2020-right}
Schwartz, R., Stanovsky, G., Swayamdipta, S., Dodge, J., and Smith, N.~A.
\newblock The right tool for the job: Matching model and instance complexities.
\newblock In \emph{Proceedings of the 58th Annual Meeting of the Association
  for Computational Linguistics}, pp.\  6640--6651, Online, July 2020.
  Association for Computational Linguistics.

\bibitem[Wolf et~al.(2019)Wolf, Debut, Sanh, Chaumond, Delangue, Moi, Cistac,
  Rault, Louf, Funtowicz, and Brew]{Wolf2019HuggingFacesTS}
Wolf, T., Debut, L., Sanh, V., Chaumond, J., Delangue, C., Moi, A., Cistac, P.,
  Rault, T., Louf, R., Funtowicz, M., and Brew, J.
\newblock Huggingface's transformers: State-of-the-art natural language
  processing.
\newblock \emph{ArXiv}, abs/1910.03771, 2019.

\bibitem[Yang et~al.(2019)Yang, Dai, Yang, Carbonell, Salakhutdinov, and
  Le]{NIPS2019_8812}
Yang, Z., Dai, Z., Yang, Y., Carbonell, J., Salakhutdinov, R.~R., and Le, Q.~V.
\newblock Xlnet: Generalized autoregressive pretraining for language
  understanding.
\newblock In \emph{Advances in Neural Information Processing Systems 32}, pp.\
  5753--5763. Curran Associates, Inc., 2019.

\end{thebibliography}
\bibliographystyle{icml2019}

\end{document}